\documentclass[runningheads]{llncs}
\usepackage{nopageno}
\usepackage{graphicx} 
\usepackage{url}
\usepackage[parfill]{parskip}
\usepackage{graphicx}
\usepackage{subcaption}
\newcommand{\printfnsymbol}[1]{%
  \textsuperscript{\@{*}}%
}

\title{Automating question generation from educational text}

\date{September 2023}

\begin{document}

\author{Ayan Kumar Bhowmick \and Ashish Jagmohan \and Aditya Vempaty \and Prasenjit Dey \and Leigh Hall \and Jeremy Hartman \and Ravi Kokku \and Hema Maheshwari}
\institute{Merlyn Mind Inc.
}

\maketitle

\begin{abstract}
The use of question-based activities (QBAs) is wide-spread in education, traditionally forming an integral part of the learning and assessment process. In this paper, we design and evaluate an automated question generation tool for formative and summative assessment in schools. We present an expert survey of one hundred and four teachers, demonstrating the need for automated generation of QBAs, as a tool that can significantly reduce the workload of teachers and facilitate personalized learning experiences. Leveraging the recent advancements in generative AI, we then present a modular framework employing transformer based language models for automatic generation of multiple-choice questions (MCQs) from textual content. The presented solution, with distinct modules for question generation, correct answer prediction, and distractor formulation, enables us to evaluate different language models and generation techniques. Finally, we perform an extensive quantitative and qualitative evaluation, demonstrating trade-offs in the use of different techniques and models.
\end{abstract}

\section{Introduction}
Recent advancements in generative artificial intelligence (AI) techniques~\cite{genAI-1}, propelled by the development of large language models (LLMs)~\cite{gpt-2,gpt3}, have demonstrated remarkable capabilities in a wide array of natural-language processing tasks. State-of-the-art models such as ChatGPT and GPT-4~\cite{gpt4} have shown promising results in tasks like text summarization, translation, and question answering, setting a new benchmark for natural language understanding. These breakthroughs have sparked a growing interest in leveraging the potential of generative AI to address challenges and automate tasks in several domains. In particular, education is expected to be one of the sectors most impacted by generative AI\footnote{\url{https://tinyurl.com/97e7bwv3}}.

Generative AI has several applications in education, such as developing conversational agents to analyze student data and interactions for personalized feedback and guiding students through learning materials. It can also help automate time-consuming administrative tasks for teachers and analyze individual student performance for providing personalized feedback on necessary improvements, thereby saving teachers a significant amount of time and allowing them to maximize their focus on in-classroom teaching. While generative AI should not replace teachers, it can assist them in augmenting their abilities to enhance the learning experience for students.

Automated question generation for assessments in the form of question-based activities (QBAs)~\cite{aie1,aie} holds significant importance in modern classrooms as it alleviates teachers' burden, allowing for more focus on student-centered endeavors. AI-driven techniques for automated question generation can greatly simplify the process of generating meaningful and relevant questions from textual educational material. This would facilitate personalized and engaging learning experiences, efficient evaluation of students' understanding, targeted feedback, and improved educational outcomes for students.

In this paper, we first conduct a survey of teachers to understand the importance of QBAs  and the challenges faced in preparing them. Based on the survey insights, we leverage recent generative AI advancements, we develop a system for automatically generating multiple-choice questions (MCQs) from educational text. We follow a modular approach for automatic generation of MCQs with separate modules for question generation, correct answer pre-
diction and distractor formulation, implemented using transformer based language models such as T5 and GPT-3. Our goal is to create a scalable, reliable, and privacy-preserving question generation framework, prioritizing these aspects over high accuracy for the educational domain. The modular approach enhances the proposed system's adaptability and efficiency, catering to various educational requirements. Compared to state-of-the-art LLMs such as GPT-4 which could generate MCQs of superior quality, smaller models such as T5 and GPT-3 may trade off MCQ quality. Nevertheless, teachers can always refine the generated MCQs to better suit their needs but concerns related to latency, privacy and reliability (GPT-4 suffers from these) remain
vital and non-negotiable in the field of education.

The proposed modular framework for MCQ generation offers various benefits, including independent development and optimization of each module and integration of diverse language models. This helps to accommodate the strengths and minimize the limitations of individual models in generating various MCQ components, while making the framework adaptable and future-proof. This flexibility also enhances the overall MCQ quality and achieves a wide range of educational objectives. Thus, the proposed framework successfully combines the advantages of T5, GPT-3, and other transformer based language models, resulting in a robust, scalable, and highly customizable MCQ generation system.

Our quantitative evaluation shows that the proposed modular MCQ generation framework generates high quality question text with decent grammar and well-formedness. It also has reasonably good accuracy in predicting correct answers. Qualitative evaluation conducted by human annotators further substantiates the quality of questions, answers, and effectiveness of the generated distractors. The annotators found that distractors were indistinguishable from correct answers, making it challenging to guess the correct option based on MCQ structure.

In short, major contributions of this paper include: (a) conducting a survey of teachers to understand the need for an AI-assisted question generation system for QBAs to help teachers, (b) proposing an end-to-end AI Question Generation (\emph{QGen}) system for generating relevant and content grounded MCQs from educational text, (c) creating a flexible framework with individual modules accommodating different generative models for generating questions, answers and distractors, and (d) evaluating the \emph{QGen} system both quantitatively and qualitatively, demonstrating its effectiveness in creating high quality MCQs.

\section{Question-based Activities in Education} \label{survey}
There is a wealth of literature that explores how formative and summative assessments can improve learning outcomes by measuring learner progress along various objectives and identifying gaps~\cite{bhat2019formative,stanja2023formative}. While assessments can take a variety of forms (essays, discussions, surveys etc.), question-based activities (QBAs) have traditionally been an integral part of assessment and the learning process. QBA’s are all activities that invite students to answer questions, which can include unit tests, quizzes, games, exit tickets, study or practice sessions, small group discussions and worksheets. These activities may be used to help students learn concepts and skills, as well as to assess their knowledge and abilities.   

In order to better understand the role of QBAs in the modern classroom, we conducted a survey on Survey Monkey that examined how teachers use QBAs in their instruction. One hundred and four teachers in grades three through eight participated. All teachers taught English language arts, science, and/or social studies. The survey included a range of queries examining: (i) How frequently do teachers use QBAs, (b) what outcomes are teachers trying to achieve by using QBAs, (iii) What sources, materials and tools do they use to create their QBAs? (iv) How much time do teachers spend creating QBAs, and (v) what challenges do teachers experience with QBAs?

Respondents answered 33 questions that were a mix of multiple-choice and written responses. The written responses were analyzed following the recommendations in~\cite{huberman2014qualitative,saldana2016coding}. We first analyzed each question individually. Responses were read repeatedly in order to identify patterns within them.  We looked across pattern codes to determine what assertions could be made about each of the areas we were investigating. Once assertions had been identified, we reread the pattern codes and determined which assertion, if any, they served as evidence for. Pattern codes were then grouped under the appropriate assertion.
The survey resulted in several interesting findings, that are summarized below.

\begin{figure} [thb]
\begin{subfigure}{.35\textwidth}
  \centering
  \includegraphics[width=1.1\linewidth]{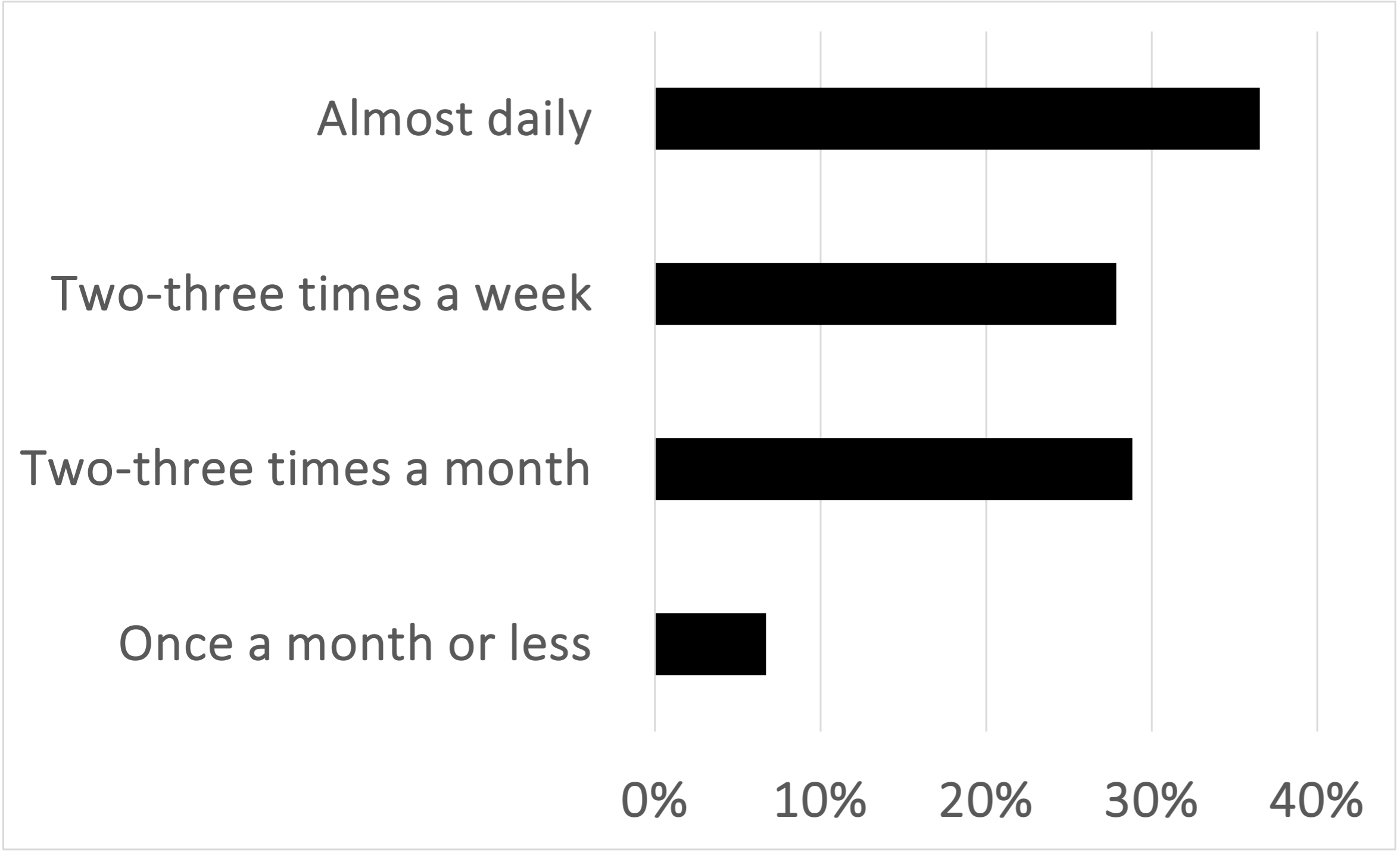}
  \caption{How frequently do teachers use QBAs in instruction?}
  \label{fig:req}
\end{subfigure} \hspace{0.1\textwidth}
\begin{subfigure}{.4\textwidth}
  \centering
  \includegraphics[width=1.1\linewidth]{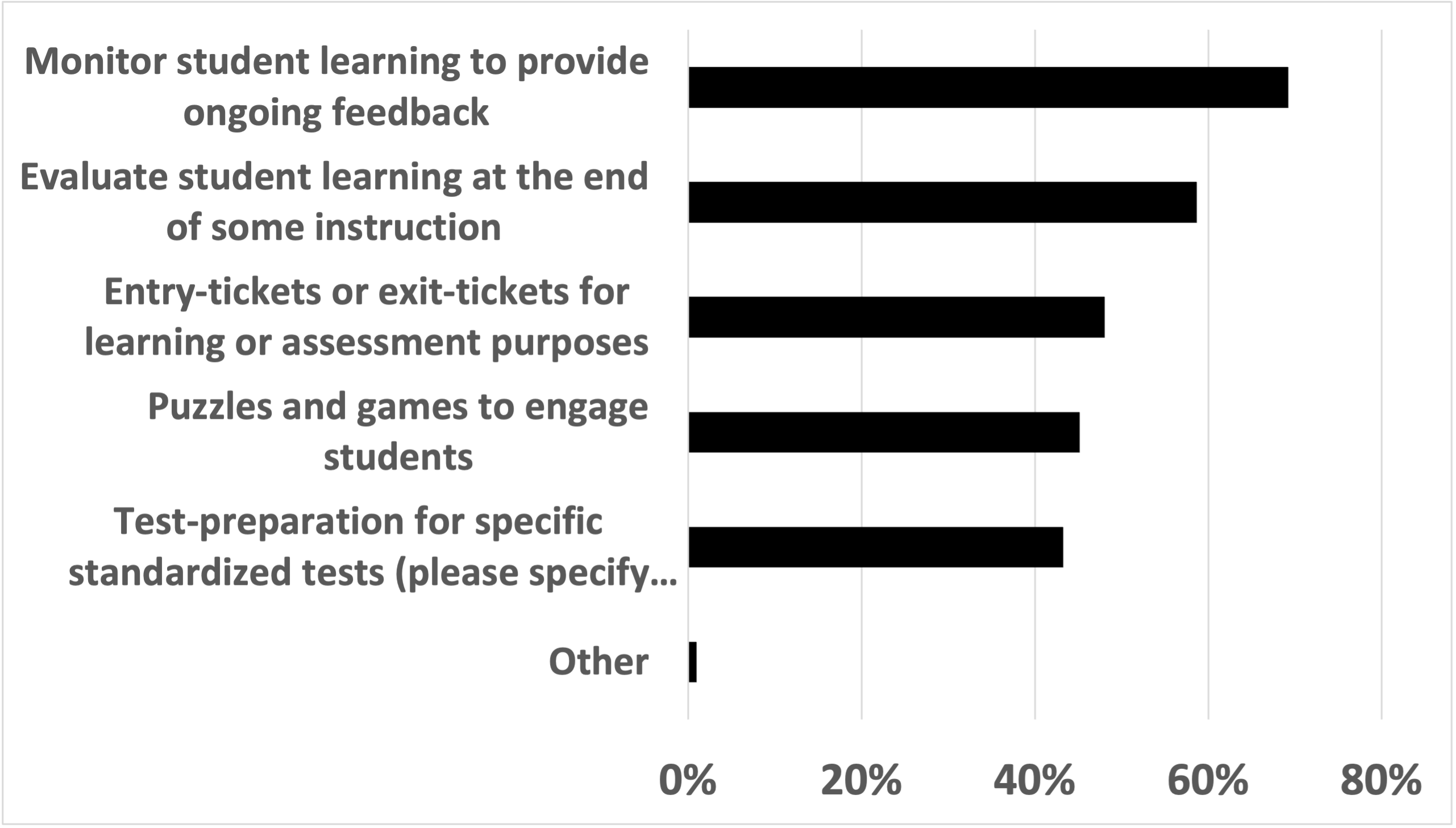}
  \caption{What outcomes are teachers trying to achieve?}
  \label{fig:ocome}
\end{subfigure}
\caption{Frequency of QBA usage in classroom instruction, and desired outcomes}
\label{fig:outcome}
\end{figure}

\textbf{Frequency of usage and outcomes}: Figure \ref{fig:outcome} confirms the ubiquity of QBAs in instruction. 28\% teachers reported using QBAs two-three times a week, and 37\% reported using them daily. While teachers’ purposes for using QBAs varied, 70\% were using them to monitor student learning and provide ongoing feedback while 59\% were using them as a way to evaluate student learning after a set period of instruction. Other uses of QBAs included: (a) entry and/or exit tickets (used by 48\% of teachers), (b) creating puzzles and games to engage students (45\% of teachers) and (c)standardized test preparation (43\%). In summary, our data suggests that QBA usage is widespread in the classroom and that teachers use them to achieve a variety of objectives.

\begin{figure} [t]
\begin{subfigure}{.35\textwidth}
  \centering
  \includegraphics[width=1.1\linewidth]{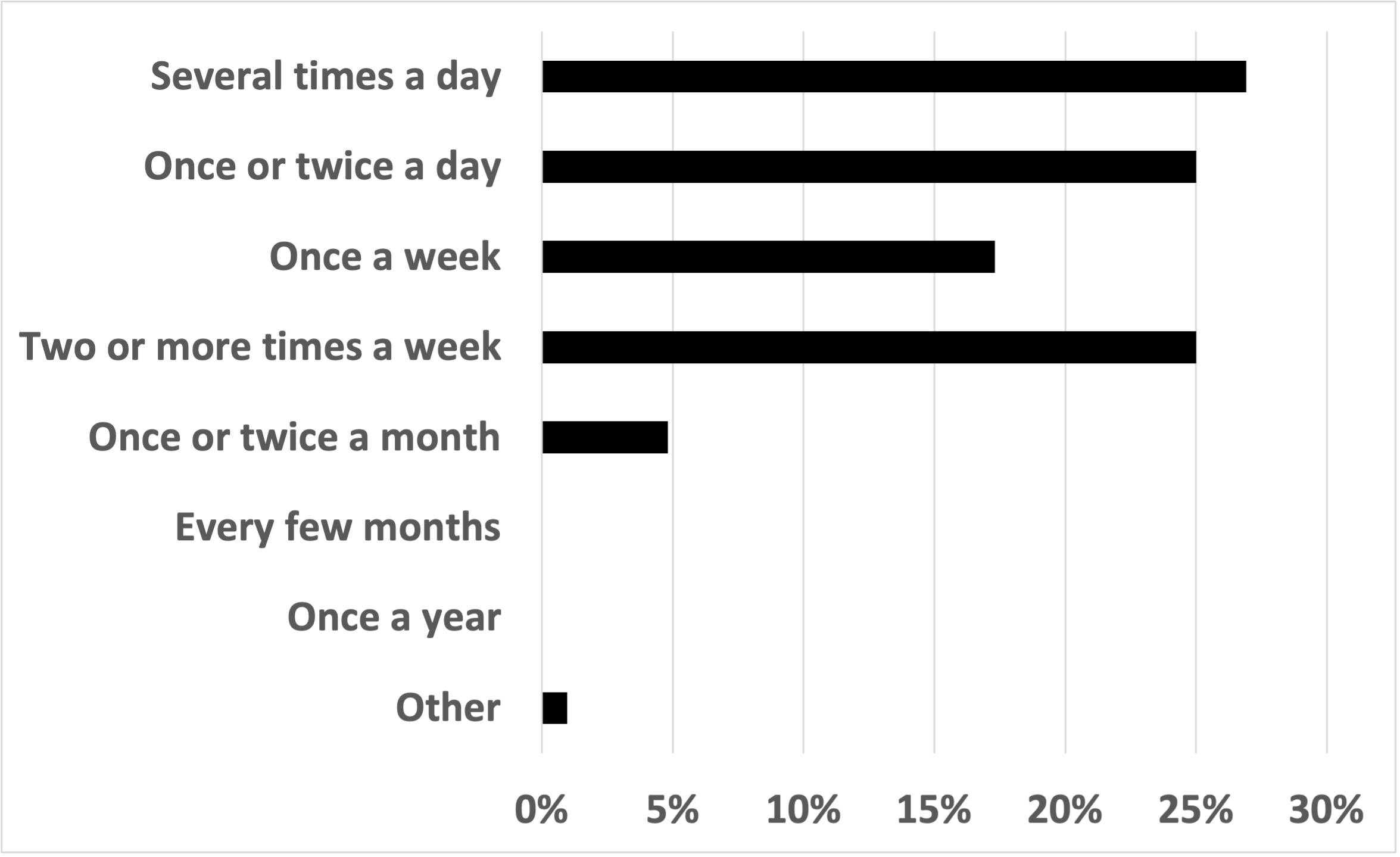}
  \caption{How frequently do teachers prepare QBAs?}
  \label{fig:req}
\end{subfigure} \hspace{0.1\textwidth}
\begin{subfigure}{.35\textwidth}
  \centering
  \includegraphics[width=1.1\linewidth]{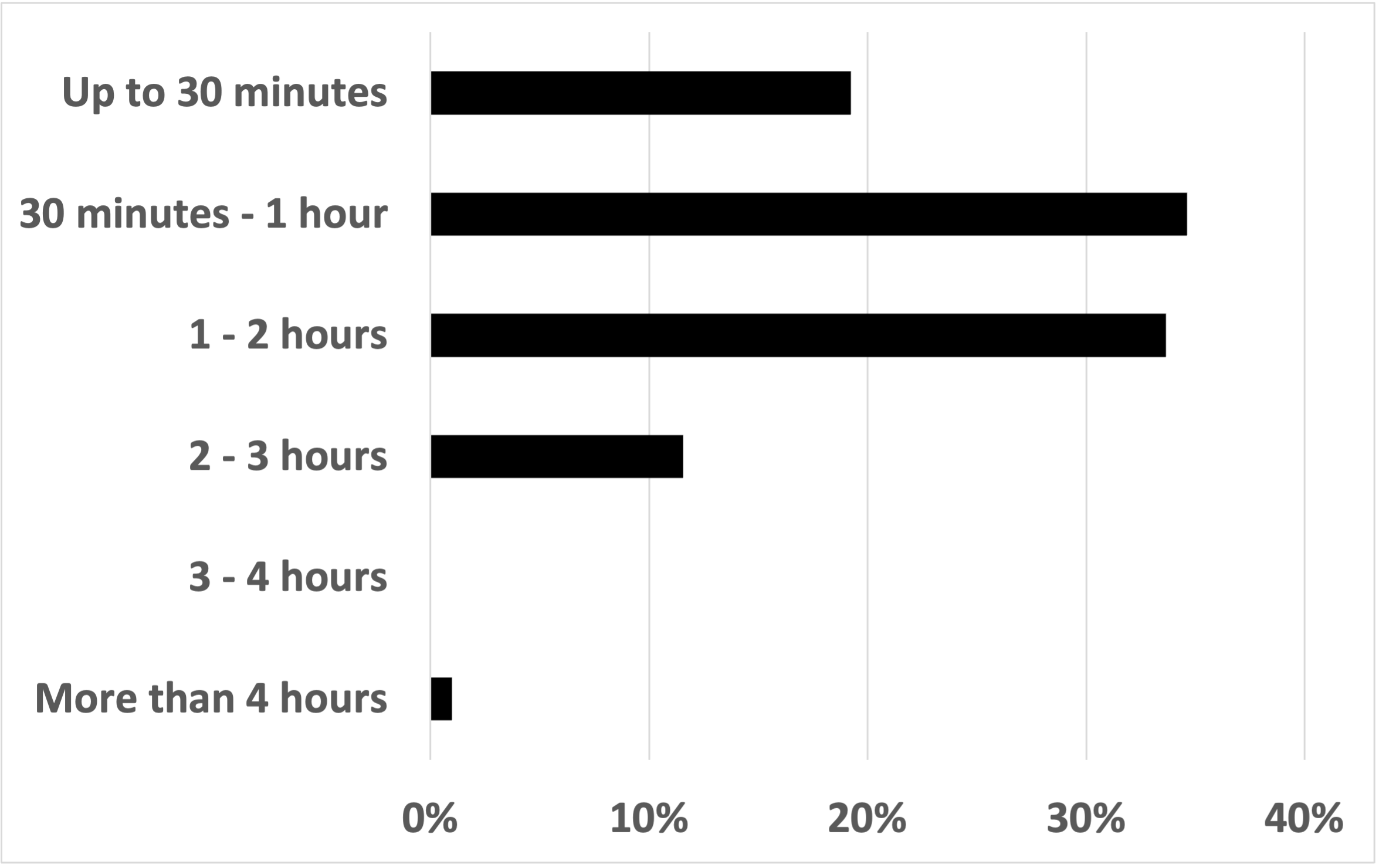}
  \caption{How much time is spent on each prep session?}
  \label{fig:ocome}
\end{subfigure}
\caption{Frequency and time spent on QBA preparation.}
\label{fig:time}
\end{figure}

\textbf{Creation time}: Figure \ref{fig:time} shows how, despite (or perhaps because of) the diversity of available sources, teachers still spend considerable amount of time on preparing QBAs. First, 52\% reported spending time on preparing QBAs at least daily. Out of that, 27\% reported working on QBAs multiple times daily. Additionally, 25\% reported working on QBAs multiple times a week. Further, 69\% of teachers reported spending 30 minutes to two hours per session of QBA preparation. This suggests that most teachers allocate multiple days a week, at a minimum of 30 minutes per session, to creating QBAs. This can result in teachers spending a minimum of 2.5 hours per week on creating QBAs.

\begin{figure} [t]
\centering
  \includegraphics[width=0.46\linewidth]{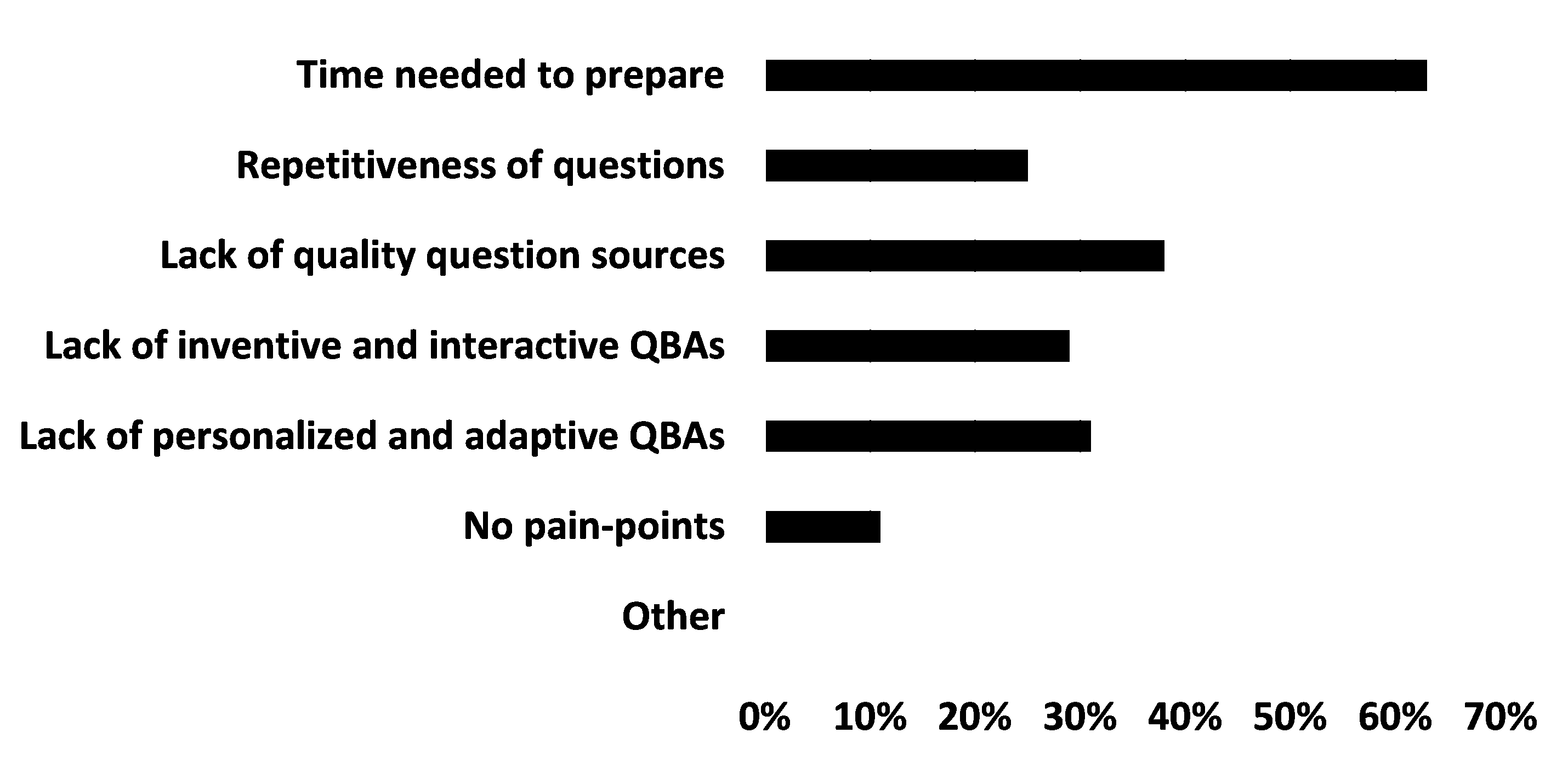}
  \caption{Key challenges faced by teachers in creating QBAs.}
  \label{fig:painpoint}
\end{figure}

\textbf{Challenges}: As Figure \ref{fig:painpoint} shows, teachers still experienced challenges in preparing QBAs despite the amount of time they spent on them. 63\% noted they did not have enough time to prepare questions for class, making preparation time the single most significant reported challenge. 38\% noted that they did not have access to quality question preparing resources, and 31\% said there was a lack of personalized and adaptive question-asking resources that would work for each class.

In short, we can summarize the main findings of the survey as follows: (i) QBA usage remains ubiquitous for instruction, (ii) Teachers spend significant amounts of time on preparing QBAs, and (iii) time, sources and question adaptivity remain challenging. Given that AI can potentially help with mitigating all of these, the survey included several queries on AI use, which yielded another important set of insights. 

\textbf{AI use}: Most surveyed teachers (82\%) had not used an AI tool to generate QBAs in the past. Of those who had, most had used tools that retrieved questions from question-banks (rather than create questions from content) while more than 50\% of such teachers reported that such tools saved time. Interestingly, almost 60\% of teachers had, currently or in the past, had human assistants, and 70\% of such teachers used their human assistants to create QBAs. 

\textbf{Requirements and concerns}: Most teachers ($>70\%$) reported interest in using AI tools to generate questions for QBAs, believing that such tools could save time, generate real-time questions that would seamlessly blend with their instruction, and improve question diversity and personalization. Teachers expressed strong preferences in having such tools integrated with lesson content and existing digital tools.

Based on the above, we believe that AI generated QBAs can fill an important gap for teachers, freeing them up for other instructional tasks, and teachers are open to such AI assistance. In the remainder of this paper, we will describe the AI question-generation system we have built, based on the insights above.

\section{Solution Architecture} \label{model}
\begin{figure} [t]
\centering
  \includegraphics[width=0.84\linewidth]{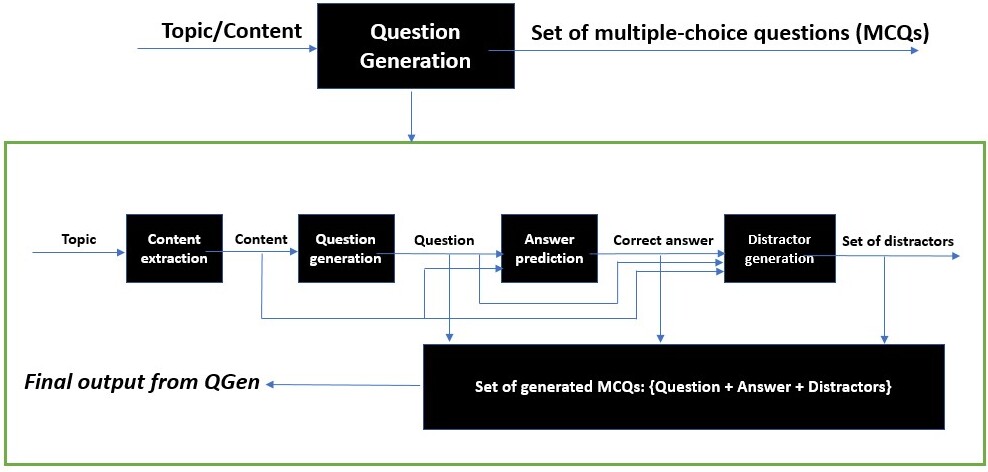}
  \caption{Architectural flow of the proposed Question Generation system (QGen)}
  \label{qgen-archit}
\end{figure}

In this section, we describe the proposed Question Generation system. The architecture of our proposed AI \textbf{Q}uestion \textbf{Gen}eration system (henceforth, we will refer to it as \emph{QGen}) is shown in Figure~\ref{qgen-archit}. As shown in the figure, \emph{QGen} takes either a topic (eg. ''American Civil War``, ''Mahatma Gandhi`` etc.) or a content (we use the terms 'context` and 'content` interchangeably throughout the paper), which is usually a body of text extracted from a given source, (eg. a summary paragraph from Wikipedia page of ''Mahatma Gandhi`` or a text snippet from the corresponding page of ''Mahatma Gandhi`` in \url{history.com}) as input modalities. \emph{QGen} generates a set of multiple-choice questions, grounded in the input context, and returns them as output. Each generated multiple-choice question (MCQ) has three components - the question text, the correct answer text (that forms one of the $n$ options of the MCQ) and a set of $n-1$ distractors that forms the remaining options of the MCQ. The correct answer and the set of distractors might appear in any random order in the generated MCQ. \emph{QGen} comprises of five modules to generate MCQs, described in subsequent sections.  

\subsection{Module 0: Content extraction from topic}
This module takes a topic as input and extracts a piece of text which may range from a few sentences to a paragraph or a sequence of multiple paragraphs retrieved from a given source (eg. school repository, Wikipedia, \url{history.com} etc.). This extracted text is then returned as the context to be used as input for question generation. Note that this step is ignored if topic is not provided as input by the end user.  

\subsection{Module 1: Question generation from content}
This module takes the context (either provided by the end user or generated from some topic in Module 0) as input and generates a set of natural-language questions relevant and grounded to the context.  
We perform both quantitative and qualitative evaluation (based on human judgement) of the quality of generated questions.
We use the following metrics to quantitatively evaluate the quality of generated questions:

\textbf{a. Perplexity:} Perplexity (PPL) is a standard metric used for evaluating the quality of text generated using language models~\cite{lm-eval}. 
We compute the perplexity of each question text generated in Module 1 of \emph{QGen} to evaluate their quality. In general, lower the perplexity value of a question, better is its quality.

\textbf{b. Query well-formedness:} This metric evaluates the well-formedness score of a generated question~\cite{wf1} i.e. if the question is incomplete or grammatically correct. The value of query well-formedness varies in the range of $[0,1]$. Higher the value of well-formedness, better is the question quality.

\subsection{Module 2: Correct answer prediction from content for a generated question}
For each question generated in Module 1, we predict the correct answer in Module 2 by identifying the answer span within the content from which the question was generated. This task is similar to a closed domain factual question answering system~\cite{nq}.
We perform both quantitative and qualitative evaluation (based on human judgement) to determine the correctness of the predicted answer to a generated question.
We rely on the following metrics to quantitatively evaluate the correctness of the predicted answer given the reference answer string:

\textbf{a. Exact Match:} Exact Match (EM)~\cite{eval-metric} is a binary metric whose value is equal to $1$ if the predicted correct answer string is exactly the same as the reference correct answer string, otherwise its value is equal to $0$.

\textbf{b. F1-score:} This metric denotes the extent of word overlap~\cite{eval-metric} between predicted and reference correct answer strings with value in the range of $[0,1]$.  

\textbf{c. ROUGE:} This is a suite of metrics to evaluate the quality of the predicted correct answer string by measuring its correspondence with a reference correct answer string~\cite{rouge1} in terms of unigram, bigram and n-gram overlaps. The metric values range between $[0,1]$. 

\subsection{Module 3: Distractor generation}
This is the last module of \emph{QGen} that corresponds to the generation of the set of distractors for a given question generated in Module 1. Distractors are options in a MCQ~\cite{dist1} other then the correct answer option (obtained in Module 2). Ideally, distractors should be as conceptually close to the correct answer as possible so that it becomes difficult to guess the correct answer from among the set of options in the MCQ. Thus, Module 3 takes the generated question, the corresponding predicted answer as well as the content from which the question-answer pair has been generated as input. It generates a set of distractors that are similar in type to the predicted (correct) answer 
for the question, though not exactly the same as the correct answer.   
To evaluate the quality of generated distractors for a MCQ, we rely only on qualitative evaluation wherein we use human judgement to determine if the set of generated distractors are acceptable, given the generated question and the predicted correct answer or if they are poor (one or more distractors deviate in type from the correct answer, making it trivial to guess the correct MCQ option).

\subsection{Filtering module} \label{filter}
In this section, we use the module specific evaluation metrics defined in previous sections in order to filter out the poorly generated MCQs and the reduced set of MCQs after filtering is the final output of \emph{QGen} provided to the end user. We consider a generated MCQ as a poorly generated candidate and filter it out if one or more of the following conditions are satisfied:

\textbf{a. Generated question is of bad quality:} Question is either grammatically incorrect, not naturally sounding, not well-structured or it is an incomplete question (reflected by high perplexity and/or low query-wellformedness value).

\textbf{b. Generated question is not grounded in the content}: The question cannot be answered correctly using the input content (Module 2 predicts correct answer to the question with a pretty low confidence score). 

\textbf{c. Predicted answer is incorrect:} Answer predicted by Module 2 for a generated question is incorrect (reflected by low confidence score).

\textbf{d. Predicted answer is same as topic}: For a generated MCQ, the correct answer predicted is same as the topic input by user from which the content was extracted in Module 0.

\textbf{e. Predicted answer is a span within the corresponding generated question}: The correct answer predicted is contained within the question text itself for a generated MCQ.

\textbf{f. Generated distractors are poor}: The set of distractors conceptually belong to a different type of entity than the correct answer entity.

\textbf{g. Repetitiveness of MCQs}: If the same MCQ is generated multiple times from given content, we prefer to keep only one occurrence of that MCQ and delete all other occurrences to ensure good diversity in the final generated set.

\textbf{h. MCQ has more than one correct answer}: Multiple options correspond to correct answers for the generated MCQ.
\section{Evaluation and results}
In this section, we evaluate the performance of \emph{QGen} in  
terms of generating high quality MCQs for a given textual content. Firstly, we introduce the datasets we use for analyzing the performance of individual \emph{QGen} modules. Next, for each individual module, we specify the model choices we have employed for performing the corresponding modular tasks. We then present the results corresponding to both quantitative and qualitative evaluation. Finally, we summarize the different variants of \emph{QGen} we have experimented with using combinations of various model choices for individual modules and compare their performance. 
\subsection{Datasets} \label{data}
For evaluation of \emph{QGen} modules, we rely on three datasets. One of them is the reading comprehension dataset \emph{SQuAD2.0}, consisting of questions posed by crowdworkers on a set of Wikipedia articles\footnote{\url{https://rajpurkar.github.io/SQuAD-explorer/explore/v2.0/dev/}}. Another dataset consists of 869 human generated MCQs (with 4 options) limited to science, ELA and social sciences domain,  extracted from a learning platform~\cite{onlinePlatform}. Thirdly, we rely on a set of 100 topics from Wikipedia~\cite{wiki} and the data consists of corresponding reading passages extracted from Wikipedia pages.
\subsection{Model details}
We rely on the following model choices for individual modules of \emph{QGen}:
\begin{enumerate}
    \item \textbf{Question generation in Module 1:} We generate the set of questions from a given piece of content using T5-based transformer models~\cite{t5} 
    as well as the recently released InstructGPT model by OpenAI. 
    We use one of two different sized T5 models, namely a fined-tuned version of T5-base model
and a fine-tuned version of T5-large model, 
both of which are trained on the \emph{SQuAD} dataset\footnote{\url{https://tinyurl.com/yntc6thk}, \url{https://tinyurl.com/mrymx4b8}} for generating questions based on some content. 
    
    On the other hand, \emph{Instruct GPT} is a large language model (LLM) that has been developed by aligning the GPT-3 LLM~\cite{gpt3} from OpenAI, with user intent on a wide range of tasks by supervised fine-tuning followed by reinforcement learning from human feedback (RLHF)~\cite{rlhf}.
    In case of InstructGPT, we use few-shot prompts consisting of three to five passages (content) with a set of example questions for each passage, for generating questions following a similar style and type for the input content.  
 
    \item \textbf{Answer prediction in Module 2:} For each question, given the content, we use two different models to predict the (correct) 
answer (usually a span within the content) to the generated question, depending on the type of model used in Module 1. For instance, if T5 is used for question generation in Module 1, we use the \emph{RoBERTa} model~\cite{roberta} trained on \emph{SQuAD2.0} dataset (described in previous section)\footnote{\url{https://github.com/facebookresearch/fairseq/tree/main/examples/roberta}}, which is an improvement over the \emph{BERT} model~\cite{bert} leading to better downstream task
performance. 
\begin{table}[t]
    \centering
    \scalebox{0.6}{
    \begin{tabular}{|c|c|c|c|}
    \hline
         \textbf{Type of QGen model used} & \textbf{Module 1} & \textbf{Module 2} & \textbf{Module 3}\\
         \hline
        \textbf{T5-base} &  \emph{T5-base} & \emph{RoBERTa-large} & \emph{Ensemble}\\
        \textbf{T5-large} &  \emph{T5-large} & \emph{RoBERTa-large} & \emph{Ensemble}\\
        \textbf{Fixed prompt GPT} & \emph{Instruct GPT} & \emph{Instruct GPT} & \emph{Instruct GPT}\\
        \textbf{Variable prompt GPT} & \emph{Instruct GPT} & \emph{Instruct GPT} & \emph{Instruct GPT}\\
        \textbf{Hybrid} & \emph{T5-large} & \emph{RoBERTa-large} & \emph{Instruct GPT}\\
        \hline
    \end{tabular}
    }
    \caption{Model configuration of individual modules for different \emph{QGen} variants}
    \label{qgen-type}
\end{table} 

On the the other hand, we use \emph{InstructGPT} model for answer prediction if \emph{InstructGPT} has been used to generate questions in Module 1. 
Similar to Module 1, we use few-shot prompts for \emph{InstructGPT} consisting of three to five passages with a set of example questions and also the corresponding answers for each passage. Then we predict the correct answer based on this prompt for the input content and a generated question from the content.

    \item \textbf{Distractor generation in Module 3:} For each pair of generated question and 
predicted answer, given the content, we rely on one of two methods depending on the variant of \emph{QGen} model (determined by model choices taken in previous modules) to generate distractors. For instance, we use \emph{InstructGPT} to generate distractors if it is used as the model in Modules 1 and 2 as well.  
In that case, similar to Module 2, we use few-shot prompts but additionally, we also provide a set of three distractors along with the correct answer for each of the example questions corresponding to a passage. Then for the input triplet of content, a generated question from the content and the corresponding answer, we obtain the generated distractors as output. 

Alternatively, if T5 and \emph{RoBERTa} are chosen as the models in Modules 1 and 2, we use an ensemble approach which is a  
combination of methods based on sense2vec, wordnet, conceptnet, 
densephrases\footnote{\url{https://tinyurl.com/fevvjuzw}, \url{https://tinyurl.com/y4p9n4wc}} as 
well as human curated MCQ datasets to generate relevant distractors. 
For \emph{sense2vec}, we generate candidate distractors that are similar to correct answer entity in terms of sense2vec embeddings. In case of both \emph{wordnet} and \emph{conceptnet}\footnote{\url{https://wordnet.princeton.edu/}, \url{https://conceptnet.io/}}, we return co-hyponyms of the hypernym of correct answer entity as the set of relevant distractors. 

In case of \emph{Densephrases}~\cite{dpr2}, we retrieve the top-$k$ most relevant passages useful for answering the generated question provided as input. Then we retrieve a list of entities from each top-$k$ passage and rank the combined list in decreasing order of semantic similarity with correct answer. Top-$n$ entities from this final ranked list are chosen as distractors for corresponding MCQ. Finally, we rely on community driven free MCQ datasets such as the open source Trivia API 
and the SciQ dataset\footnote{\url{https://the-trivia-api.com/}, \url{https://allenai.org/data/sciq}} 
for generating distractors for each unique answer option across the ~26K MCQs in these two datasets.\footnote{We use these datasets only to evaluate our research 
for non-commercial purposes.}   
\end{enumerate}

Based on these model choices, we obtain different \emph{QGen} variants depending on the model type used in each individual module, as summarized in Table~\ref{qgen-type}.
\subsection{Evaluation results}
\subsubsection{Quantitative evaluation}
Here we quantify the performance of the Modules 1, 2 and 3 for the QGen system in terms of the evaluation metrics defined in Section~\ref{model} as follows: 

\textbf{Module 1: } We evaluate Module 1 by considering the set of reading passages (or content) available in \emph{SQuAD2.0} dataset. We provide each such content as input to Module 1 of \emph{QGen} in order to generate a set of grounded and relevant questions. We repeat this exercise across each content and obtain a combined set of generated questions across all the reading passages. Finally, we compute the value of perplexity (using the causal language model GPT-2~\cite{gpt2}) and well-formedness score for each generated question. We obtain a mean value of perplexity equal to $37.3$ and a mean value of query well-formedness score equal to $0.864$ over all the generated questions. This verifies the high quality of generated questions obtained from Module 1 of \emph{QGen} in terms of natural soundedness, completeness and grammatical correctness. 

\textbf{Module 2: } Next, we evaluate Module 2 by providing each question available in \emph{SQuAD2.0} dataset and its corresponding reading passage as input to Module 2 of \emph{QGen}. We then obtain a predicted answer for each such question as the output of Module 2. We now compare the predicted answer string with the reference answer to the question available in \emph{SQuAD2.0} dataset.
We compute the values of \emph{EM}, \emph{F1-score} and \emph{ROUGE} for each predicted answer and compute their mean values over all question-answer pairs.    
We obtain a mean \emph{EM} score of $0.64$ which means that the predicted answer string exactly matches the reference \emph{SQuAD2.0} answers for $64\%$ of cases. Similarly, the mean \emph{F1-score} obtained is $0.84$ which is consistent with the \emph{EM} values as \emph{F1-score} considers the word overlap and is less stricter than \emph{EM} score. Further, mean \emph{ROUGE} scores are also pretty high with \emph{ROUGE-1 = 0.84}, \emph{ROUGE-2 
= 0.54} and \emph{ROUGE-L = 0.84}. This denotes a high overlap of word units (n-grams) between the predicted and reference \emph{SQuAD2.0} answers. Overall, we can say that Module 2 of \emph{QGen} performs reasonably in terms of predicting the correct answer to questions. 

\textbf{Module 3:} We use the dataset from the learning platform (see Section~\ref{data}) to evaluate the quality of generated distractors in Module 3. Precisely, we consider the question text as well as the correct answer option for each MCQ in this dataset as input to Module 3 and generate a set of distractors as output. We then determine if the set of distractors are acceptable or not to humans. If the distractors are topically unrelated to the correct answer or if multiple distractors are different to the correct answer in terms of entity type or semantic similarity, we consider the generated distractor set for the MCQ to be unacceptable or poor. Following this criteria, we observe that $58\%$ of the MCQs in the data from learning platform have acceptable set of generated distractors. Given the difficulty of generating good and relevant distractors for MCQs, we consider the performance of Module 3 of \emph{QGen} to be pretty decent.
\begin{figure} [t]
\centering
  \includegraphics[width=0.52\linewidth]{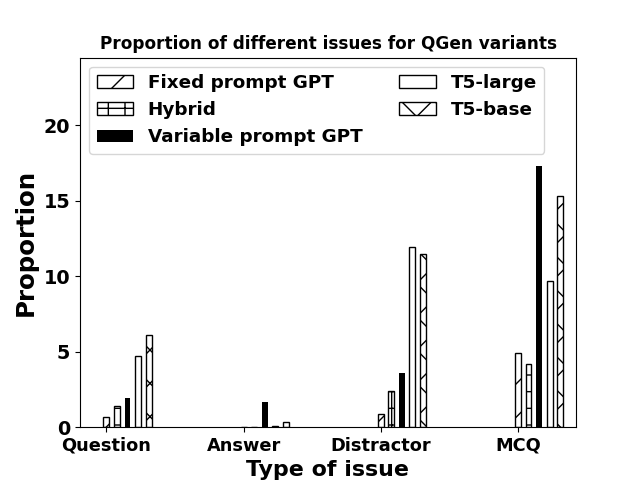}
  \caption{Different issues identified by annotators for variants of \emph{QGen}}
  \label{annot}
\end{figure}

\subsubsection{Qualitative evaluation}
In this section, we compare the quality of the MCQs generated by the different variants of \emph{QGen} shown in Table~\ref{qgen-type} through human judgement. For this purpose, we generated a set of $500$ MCQs from each \emph{QGen} variant from the set of $100$ reading passages in the \emph{Wikipedia} dataset (see Section~\ref{data}). We then hired a group of three annotators to evaluate each generated MCQ and label them with one or more of the following types of issues (see Section~\ref{filter}) if they exist with the MCQ: 
\begin{enumerate}
    \item \textbf{Question:} 
    The corresponding question in the MCQ is of bad quality.
    \item \textbf{Answer:} The predicted correct answer for the MCQ is incorrect.
    \item \textbf{Distractor:} The corresponding set of distractors for the MCQ are of poor quality or there are multiple correct answers for the MCQ.
    \item \textbf{MCQ:} The MCQ as a whole has an issue. This include issues such as repetitiveness, question is not grounded, the answer is trivial or same as topic, incorrect extrapolation etc.
\end{enumerate}

If the end-to-end MCQ is of good quality with none of the above four issues as a whole or in any of its individual components, we consider it to be a MCQ of desirable quality (human likeness).
From the annotations of generated MCQs, we observe that among the different \emph{QGen} variants, the \emph{Hybrid} variant where Module 1 is based on \emph{T5-large}, Module 2 is based on \emph{RoBERTa-large} and Module 3 is based on \emph{Instruct GPT} and the \emph{Fixed prompt GPT} variant where all modules are based on \emph{Instruct GPT} shows the best performance as $92\%$ and $93.53\%$ of the generated MCQs respectively are of desirable quality as agreed upon by all three annotators. On the other hand, proportion of generated MCQs that are of desirable quality (consensus reached among annotators) for the other three \emph{QGen} variants varies between $67-76\%$ of all MCQs. This shows that a combination of \emph{T5-based} (for question generation) and \emph{Instruct GPT} based (for all modules) \emph{QGen} model variants are best suited for generating high quality MCQs.

We plot the frequency distribution of the above four issue types for the different \emph{QGen} variants (see Table~\ref{qgen-type}) in Figure~\ref{annot} as labeled by the annotators. We observe that the most frequent issues for poorly generated MCQs in case of the \emph{Variable prompt GPT} model are the non-groundedness of questions in the content passage ($~12\%$ of cases) and there is incorrect extrapolation (or hallucination) wherein the four answer options are wrongly misinterpreted ($~10\%$ of cases).
On the other hand, one of the most frequent issues for \emph{QGen} variants that relies on T5 for question generation step is poorly generated distractors ($~12\%$ of cases) and repetitiveness of MCQs ($~5\%$ of cases). In addition, $~6.5\%$ of cases have an issue of answer being same as topic for such variants.

The main challenge that we faced while qualitatively evaluating a generated MCQ in this manner involves a rigorous iterative process that we underwent to calibrate researcher-annotator agreement. This helped us to reach a consensus on the quality of a generated MCQ between annotators and the research team as well as agree on the type of issue, if any, that may be present for a MCQ. In future, we aim to evaluate the time it takes for a teacher to use \emph{QGen} to generate questions for content provided by them and the effort needed to refine these questions through suitable experiments and surveying teachers. 
\section{Conclusion}
In conclusion, this paper highlights the potential of generative AI in addressing educational challenges, particularly in automating question-based activities (QBAs) for assessments. Leveraging transformer-based language models like T5 and GPT-3, we have designed a scalable, reliable, and privacy-preserving modular framework for multiple-choice question generation. Quantitative and qualitative evaluations verified the effectiveness of the proposed framework in generating high quality questions and answers as well as challenging and indistinguishable distractors, with GPT-3 based modules demonstrating better performance compared to their T5 counterparts. This work not only demonstrates the successful integration of various language models but also paves the way for further exploration of generative AI tools in educational applications, ultimately augmenting teachers' abilities and enhancing students' learning experiences.

\bibliographystyle{splncs04}
\bibliography{main}

\begin{thebibliography}{10}
\providecommand{\url}[1]{\texttt{#1}}
\providecommand{\urlprefix}{URL }
\providecommand{\doi}[1]{https://doi.org/#1}

\bibitem{aie1}
Ahmad, S.F., Rahmat, M.K., Mubarik, M.S., Alam, M.M., Hyder, S.I.: Artificial intelligence and its role in education. Sustainability  \textbf{13}(22),  12902 (2021)

\bibitem{aie}
Bethencourt-Aguilar, A., Castellanos-Nieves, D., Sosa-Alonso, J.J., Area-Moreira, M.: Use of generative adversarial networks (gans) in educational technology research  (2023)

\bibitem{bhat2019formative}
Bhat, B., Bhat, G.: Formative and summative evaluation techniques for improvement of learning process. European Journal of Business \& Social Sciences  \textbf{7}(5),  776--785 (2019)

\bibitem{lm-eval}
Chen, S.F., Beeferman, D., Rosenfeld, R.: Evaluation metrics for language models  (1998)

\bibitem{eval-metric}
Choi, E., He, H., Iyyer, M., Yatskar, M., Yih, W.t., Choi, Y., Liang, P., Zettlemoyer, L.: Quac: Question answering in context. arXiv preprint arXiv:1808.07036  (2018)

\bibitem{bert}
Devlin, J., Chang, M.W., Lee, K., Toutanova, K.: Bert: Pre-training of deep bidirectional transformers for language understanding. arXiv preprint arXiv:1810.04805  (2018)

\bibitem{gpt2}
Ethayarajh, K.: How contextual are contextualized word representations? comparing the geometry of bert, elmo, and gpt-2 embeddings. arXiv preprint arXiv:1909.00512  (2019)

\bibitem{wf1}
Faruqui, M., Das, D.: Identifying well-formed natural language questions. arXiv preprint arXiv:1808.09419  (2018)

\bibitem{gpt3}
Floridi, L., Chiriatti, M.: Gpt-3: Its nature, scope, limits, and consequences. Minds and Machines  \textbf{30},  681--694 (2020)

\bibitem{rlhf}
Griffith, S., Subramanian, K., Scholz, J., Isbell, C.L., Thomaz, A.L.: Policy shaping: Integrating human feedback with reinforcement learning. Advances in neural information processing systems  \textbf{26} (2013)

\bibitem{t5}
Grover, K., Kaur, K., Tiwari, K., Kumar, P.: Deep learning based question generation using t5 transformer. In: Advanced Computing: 10th International Conference, IACC 2020, Panaji, Goa, India, December 5--6, 2020, Revised Selected Papers, Part I 10. pp. 243--255. Springer (2021)

\bibitem{huberman2014qualitative}
Huberman, A., et~al.: Qualitative data analysis a methods sourcebook  (2014)

\bibitem{wiki}
Kriangchaivech, K., Wangperawong, A.: Question generation by transformers. arXiv preprint arXiv:1909.05017  (2019)

\bibitem{nq}
Kwiatkowski, T., Palomaki, J., Redfield, O., Collins, M., Parikh, A., Alberti, C., Epstein, D., Polosukhin, I., Devlin, J., Lee, K., et~al.: Natural questions: a benchmark for question answering research. Transactions of the Association for Computational Linguistics  \textbf{7},  453--466 (2019)

\bibitem{dpr2}
Lee, J., Wettig, A., Chen, D.: Phrase retrieval learns passage retrieval, too. arXiv preprint arXiv:2109.08133  (2021)

\bibitem{dist1}
Liang, C., Yang, X., Dave, N., Wham, D., Pursel, B., Giles, C.L.: Distractor generation for multiple choice questions using learning to rank. In: Proceedings of the thirteenth workshop on innovative use of NLP for building educational applications. pp. 284--290 (2018)

\bibitem{roberta}
Liu, Y., Ott, M., Goyal, N., Du, J., Joshi, M., Chen, D., Levy, O., Lewis, M., Zettlemoyer, L., Stoyanov, V.: Roberta: A robustly optimized bert pretraining approach. arXiv preprint arXiv:1907.11692  (2019)

\bibitem{genAI-1}
Radford, A., Metz, L., Chintala, S.: Unsupervised representation learning with deep convolutional generative adversarial networks. arXiv preprint arXiv:1511.06434  (2015)

\bibitem{gpt-2}
Radford, A., Narasimhan, K., Salimans, T., Sutskever, I., et~al.: Improving language understanding by generative pre-training  (2018)

\bibitem{saldana2016coding}
Salda{\~n}a, J.: The coding manual for qualitative researchers. kindle e-reader version (2016)

\bibitem{rouge1}
Schluter, N.: The limits of automatic summarisation according to rouge. In: Proceedings of the 15th Conference of the European Chapter of the Association for Computational Linguistics. pp. 41--45. Association for Computational Linguistics (2017)

\bibitem{onlinePlatform}
Solas, E., Sutton, F.: Incorporating digital technology in the general education classroom. Research in Social Sciences and Technology  \textbf{3}(1),  1--15 (2018)

\bibitem{stanja2023formative}
Stanja, J., Gritz, W., Krugel, J., Hoppe, A., Dannemann, S.: Formative assessment strategies for students' conceptions—the potential of learning analytics. British Journal of Educational Technology  \textbf{54}(1),  58--75 (2023)

\bibitem{gpt4}
Zhang, C., Zhang, C., Zheng, S., Qiao, Y., Li, C., Zhang, M., Dam, S.K., Thwal, C.M., Tun, Y.L., Huy, L.L., et~al.: A complete survey on generative ai (aigc): Is chatgpt from gpt-4 to gpt-5 all you need? arXiv preprint arXiv:2303.11717  (2023)

\end{thebibliography}

\end{document}